\pdfoutput=1
\documentclass[runningheads]{llncs}

\newif\ifanonymous
\anonymousfalse   %

\usepackage[T1]{fontenc}
\usepackage{graphicx}
\usepackage[numbers]{natbib}
\usepackage[linesnumbered,ruled,vlined]{algorithm2e}
\usepackage{float}
\SetKwInOut{KwInOut}{In/Out}
\usepackage{booktabs}
\usepackage{hyperref} %

\makeatletter

\renewcommand\section{\@startsection{section}{1}{\z@}%
  {-14\p@ \@plus -3\p@ \@minus -3\p@}%
  {7\p@ \@plus 2\p@ \@minus 2\p@}%
  {\normalfont\large\bfseries\boldmath
   \rightskip=\z@ \@plus 8em\pretolerance=10000 }}

\renewcommand\subsection{\@startsection{subsection}{2}{\z@}%
  {-12\p@ \@plus -3\p@ \@minus -3\p@}%
  {5\p@ \@plus 2\p@ \@minus 2\p@}%
  {\normalfont\normalsize\bfseries\boldmath
   \rightskip=\z@ \@plus 8em\pretolerance=10000 }}

\renewcommand\subsubsection{\@startsection{subsubsection}{3}{\z@}%
  {-10\p@ \@plus -2\p@ \@minus -2\p@}%
  {0.4em}%
  {\normalfont\normalsize\bfseries\boldmath}}

\makeatother

\begin{document}
\title{OpenProver: Agentic and Interactive\texorpdfstring{\\}{ } Theorem Proving with Lean 4}
\ifanonymous
  \author{Anonymous Authors}
  \authorrunning{Anonymous Authors}
  \institute{Anonymous Institution}
\else
  \author{Matěj Kripner\orcidID{0009-0006-9530-3670} \and
  Milan Straka\orcidID{0000-0003-3295-5576}}
  \authorrunning{M. Kripner and M. Straka}
  \institute{Charles University, Faculty of Mathematics and Physics, Prague, Czech Republic\\
  \email{\{kripner,straka\}@ufal.mff.cuni.cz}}
\fi

\ifanonymous
  \hypersetup{
    pdftitle={OpenProver: Agentic and Interactive Theorem Proving with Lean 4},
    pdfauthor={Anonymous}
  }
\else
  \hypersetup{
    pdftitle={OpenProver: Agentic and Interactive Theorem Proving with Lean 4},
    pdfauthor={Matěj Kripner and Milan Straka}
  }
\fi
\maketitle              %
\begin{abstract}
In this system paper, we present OpenProver, an open-source system for LLM-driven automated theorem proving (ATP) with integrated Lean 4 formal verification.
OpenProver integrates a Planner-Worker-Verifier architecture inspired by recent ATP agentic systems such as Aletheia.
A Planner agent maintains a compact Whiteboard scratchpad and an unbounded Repository of intermediate findings, and decomposes mathematical work into parallel Workers.

OpenProver is fully open-source, offers reproducible evaluation through automatic formal verification of generated proofs, and provides an interactive terminal interface for human-guided proof search.
In interactive mode, OpenProver allows the human operator to monitor and steer the proof search process, motivated by the established human-AI synergy in interactive code generation.

To showcase the potential for quantitative ablation experiments enabled by automatic formal verification, we evaluate OpenProver on ProofNet and compare it with a simple baseline.
OpenProver is publicly available at \href{https://github.com/kripner/OpenProver}{github.com/kripner/OpenProver}.

\keywords{Automated Theorem Proving \and Lean 4 \and Agentic LLMs}
\end{abstract}

\section{Introduction}

Automated Theorem Proving (ATP) has seen a significant rise in capabilities with the integration of Large Language Models (LLMs) trained using Reinforcement Learning from Verifiable Rewards (RLVR)~\citep{rlvr}.
Beyond tackling some of the hardest problems in competition mathematics, we are now seeing sparks of usefulness of ATP systems in frontier mathematical research.

Existing ATP systems can be roughly divided into two categories.
First, fully autonomous theorem provers attempt end-to-end proof generation without human intervention, enabling reproducibility across runs.
An example of such a~system is Aletheia~\citep{aletheia}.

Second, interactive theorem provers (ITPs) enable the user to monitor and intervene in the proof search process, following the observation that while autonomous AI systems do not yet match expert human performance, the integration of both can greatly accelerate research.
For example, a mathematician can formulate several potential approaches to completing a certain proof, receive worked-out counterexamples and observations from an AI assistant, and use these to iteratively adjust the high-level plan.
With a human operator in the loop, ITP approaches require careful design of the visual user interface.
Concurrent to our work, several agentic LLM-based ITP tools have been released, including OpenGauss~\citep{opengauss_repo}.

We present OpenProver, an automated theorem prover that bridges the gap between reproducible ATP research and interactive mathematical tooling.
OpenProver extends Aletheia, most notably by integrating formal verification capabilities of Lean 4~\citep{lean4}.
To showcase the automatic evaluation enabled by Lean, we present quantitative measurements of OpenProver performance on ProofNet~\citep{proofnet}, compared to a simple linear Chain-of-Thought baseline.
We hope that our work advances progress towards ATP systems that are both useful in frontier mathematical research and whose performance can be rigorously measured.

\section{System Description}

\begin{figure}
\includegraphics[width=\textwidth]{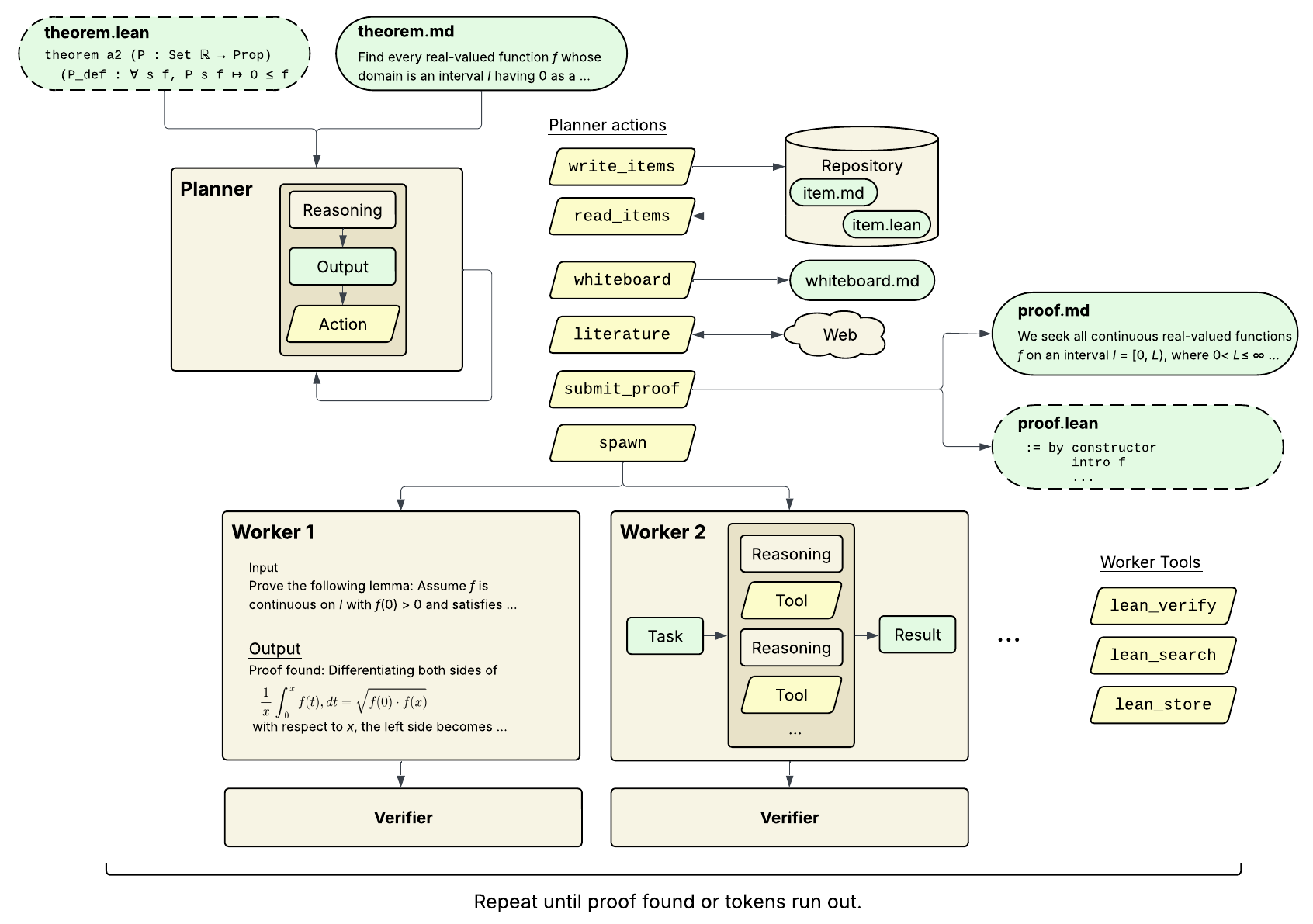}
\caption{Overview of the OpenProver proof search loop. Planner iterates on the high-level plan, manages persistent state, and delegates mathematical work to Workers. A~Verifier provides independent feedback to each Worker contribution.} \label{fig:system}
\end{figure}

\noindent
OpenProver is an automated theorem prover utilizing a reasoning LLM executed in an agentic scaffolding, with integration of the Lean 4 formal verifier.
The system operates in an iterative loop composed of three types of agents illustrated in Figure~\ref{fig:system}: a single \textbf{Planner}, parallel \textbf{Workers}, and parallel \textbf{Verifiers}.

\subsection{Architecture}

During proof search, OpenProver utilizes three distinct types of agents:

\begin{enumerate}
    \item \textbf{Planner:} Maintains global state and determines a high-level plan.
    \item \textbf{Workers:} Independently explore candidate proof strategies, lemmas, counterexamples, and similar.
    \item \textbf{Verifiers:} Independently verify each Worker output, providing additional context to the Planner.
\end{enumerate}

\noindent
The proof search process is structured as a sequence of Planner steps, described in more detail in Section~\ref{sec:proof_search_loop}.

\subsubsection{Planner}
At each step, the Planner first generates a chain-of-thought reasoning trace~\citep{cot} before producing an output and a list of actions.
The available Planner actions are listed in Table~\ref{tab:planner-actions} and explained in more detail in the following.

\begin{table}
\caption{Planner actions. One or more actions are executed in each Planner step.}
\label{tab:planner-actions}
\begin{tabular}{|l|l|l|}
\hline
\textbf{Action} & \textbf{Description} & \textbf{Input} \\
\hline
\texttt{spawn} & Spawn parallel workers & List of tasks \\
\hline
\texttt{read\_items} & Read repo items & List of slugs \\
\hline
\texttt{write\_items} & Create/update/delete repo items & List of (slug, content, format) \\
\hline
\texttt{read\_theorem} & Re-read theorem statement & none \\
\hline
\texttt{write\_whiteboard} & Update Whiteboard content & Full Whiteboard text \\
\hline
\texttt{submit\_proof} & Submit informal proof & Informal item slug \\
\hline
\texttt{submit\_lean\_proof} & Submit Lean proof & Lean item slug \\
\hline
\texttt{literature\_search}$^\ast$ & Search online literature & Query \\
\hline
\end{tabular}

\vspace{2pt}
{\footnotesize $^\ast$ Not available in isolation mode.}
\end{table}

\subsubsection{Workers}
Workers are independent agents spawned by the Planner.
For each spawned Worker, the Planner provides a plaintext task description that solely specifies the Worker's purpose.
Example tasks include advancing a specific proof direction, proposing a decomposition of a theorem into sub-goals, proving a lemma, exploring minimal cases, searching for counterexamples, or proving a simplified version of the theorem.
Additionally, Workers can be tasked with formalizing an existing natural language proof in Lean.

Importantly, a Worker does not observe the reasoning traces or outputs of previous Workers or of the Planner.
This facilitates independent exploration of distinct proof attempts, where each Worker is not swayed by irrelevant context.

\subsubsection{Verifiers}
For each finished Worker, we spawn a Verifier tasked with producing an independent natural language feedback with the goal of revealing flaws in the Worker output.
Crucially, the Verifier does not observe the Worker's reasoning trace, reducing bias toward following the same, potentially flawed, line of thinking.

\subsection{State and Memory Management}

To preserve important context between Planner steps, OpenProver manages a~single compact Markdown file called the \textit{Whiteboard}, updated periodically by the Planner using the \texttt{write\_whiteboard} action.
While the content of the Whiteboard is decided solely by the Planner, it is advised by the corresponding prompt to contain at least the currently executed proof plan, the history of all already explored and failed attempts, and ideas to return to later, together with any useful brief observations and notes.
The Whiteboard is provided to the Planner as input at each step.
This design can be seen as analogous to the Reasoning Cache~\citep{reasoning_cache}, extending the effective context length of reasoning LLMs.

Additionally, the Planner occasionally needs to store longer text or Lean snippets such as detailed failed proof attempts, proofs of lemmas, literature summaries, and similar.
To prevent overflowing the Whiteboard, OpenProver manages a~\textit{Repository} of items where each \textit{Item} is either a Markdown file or a~Lean file.
Items are organized in a folder-like structure using the \texttt{write\_items} and \texttt{read\_items} actions, and are referred to by their relative path, which we call a \textit{slug}.
At each Planner step, alongside the Whiteboard, the Planner also observes the slugs and one-line summaries of all Items stored in the Repository.
Crucially, Lean Items are only stored if they pass the Lean formal verification; otherwise, the errors and warnings are fed back to the Planner.
This enables tighter feedback from the formal verifier than just a final-answer check.

\subsection{Proof Search Loop}
\label{sec:proof_search_loop}

OpenProver executes a linear proof search loop outlined in Algorithm~\ref{alg:openprover} until either a proof is found or the compute budget runs out.
Parallelization is enabled by spawning multiple Workers.

\begin{figure}[t]
\begin{minipage}{\linewidth}

\begin{algorithm}[H]
\caption{OpenProver: high-level operation}
\label{alg:openprover-loop}
\KwIn{\textsc{theorem.md}; optionally \textsc{theorem.lean}; token budget $B$}
\KwOut{\textsc{proof.md}, \textsc{discussion.md}; optionally \textsc{proof.lean}}

Initialize Whiteboard $W \gets \emptyset$, repo $R \gets \emptyset$, history $H \gets []$\;
\While{\textnormal{budget not exhausted} \textbf{and} \textnormal{\textsc{proof.md}} \textnormal{not yet produced}}{
    \textsc{PlannerStep}($W, R, H$)\;
}
\textit{(optionally, iterate \textnormal{\textsc{PlannerStep}} further until \textnormal{\textsc{proof.lean}} is produced)}\;
write \textsc{discussion.md} summarizing proof search process and result\;
\end{algorithm}

\begin{algorithm}[H]
\caption{\textsc{PlannerStep}}
\label{alg:planner}
\KwInOut{Whiteboard $W$, repo summary $R'$, history $H$, theorem $T$}
Query planner LLM with $(W,\ R',\ H_{-n:}, T)$:\;
\Indp
  Produce reasoning trace (CoT)\;
  Produce free-form output and a list of actions $a_1,\dots,a_k$\;
\Indm
Run $a_1,\dots,a_k$ in parallel and wait for completion\;
Append each action's output to $H$\;
\end{algorithm}

\end{minipage}
\caption{High-level pseudocode of OpenProver: the main loop (Alg.~1) repeatedly invokes \textsc{PlannerStep} (Alg.~2).
}
\label{alg:openprover}
\end{figure}

\subsection{Integration with Lean}

OpenProver optionally integrates with the Lean formal verifier, with the only requirement that the formal theorem statement is provided on input in the form of a Lean file with one or more \texttt{sorry} keywords.

First, after a~natural language proof is found, OpenProver attempts to formalize it and submit the resulting Lean proof.
In case of errors or warnings emitted by the Lean verifier, OpenProver iteratively attempts to fix the proof, potentially transitioning back to fixing the natural language proof in case a flaw is identified.

Second, the Planner can verify the correctness of any intermediate result by storing it as a Lean Item in the Repository.
Third, Workers can execute Lean-related \textit{Tool Calls}~\citep{toolformer} with the following three tools available:
\begin{itemize}
\item \texttt{lean\_verify} -- Verify the correctness of a Lean snippet.
\item \texttt{lean\_search} -- Perform semantic search over Mathlib~\citep{mathlib} using LeanExplore~\citep{leanexplore}, obtaining the $k$ most relevant definitions.
\item \texttt{lean\_store} -- Append a Lean snippet to a temporary file which is then prepended to each following \texttt{lean\_verify} input. Typically, this includes imports, namespace openings, definitions, and already proven sub-lemmas.
\end{itemize}

\noindent
An essential limitation of formal ATP is the fact that formalization is often more challenging than the informal proof search.
This gap is expected to reduce as the Lean ecosystem of already proven building blocks evolves over time.

\section{Interactive User Interface}

OpenProver offers an interactive Terminal User Interface (TUI) where the user can monitor and steer the search process.
Specifically, users can:
\begin{itemize}
\item Monitor the streaming output of all agents.
\item Browse the history of previous Planner steps, inspecting their reasoning trace, action inputs, action outputs, and Whiteboard state.
\item Interrupt any Worker if its reasoning direction seems unpromising.
\item Interrupt the Planner and provide textual feedback to steer it.
\item In Manual mode, the user is prompted to accept each set of Planner actions before they are executed. When they are rejected, potentially with text feedback, the Planner adjusts before proposing new actions. This is skipped in Autonomous mode.
\end{itemize}

\section{Experiments}

\begin{table}
\caption{Performance of OpenProver versus a baseline on ProofNet across models.}
\label{tab:openprover_results}
\centering
\renewcommand{\arraystretch}{1.15}
\begin{tabular}{l@{\hspace{12pt}}c@{\hspace{12pt}}c}
\toprule
Model & Linear Rollout & OpenProver \\
\midrule
Kimi-K2.5  & 36.8\% & \textbf{57.3\%} \\
Leanstral  & 21.1\% & \textbf{28.1\%} \\
\bottomrule
\end{tabular}
\end{table}

\noindent
We measure OpenProver performance in autonomous mode on 185 formal theorems of ProofNet across different underlying models --- Kimi K2.5~\citep{kimi-k2.5} and Leanstral~\citep{leanstral} --- and a 100k-tokens per-problem budget.
Note that OpenProver is model-agnostic and can utilize any reasoning model as a building block in agentic proof search.
As a baseline, we unroll each model in a linear conversation under the same token budget, where the model can perform an unlimited number of Lean verifications.
The results of this setup are summarized in Table~\ref{tab:openprover_results}.

\section{Conclusion}

With the presentation of OpenProver, we hope to pave the way towards reproducible research in the field of agentic theorem proving through quantitative evaluations enabled by automatic formal verification.
Additionally, it is conceivable that this automatic evaluation feedback could be used as a guiding signal for an autonomous self-improvement process in the space of code or LLM prompts, akin to e.g. Feedback Descent~\citep{feedback_descent} or AlphaEvolve~\citep{alphaevolve}.
At a high level, OpenProver adopts a~flexible design that shifts many of the behavior decisions from code to prompts, making pure prompt-based self-improvement viable.

\begin{credits}
\subsubsection{\ackname}

This project was supported by the grant no. 25-18031S of the Czech Science Foundation (GAČR), by the Charles University Grant Agency (GAUK), project no. 458326, and by the CEDMO 2.0 NPO project.
This research was partially supported by SVV project number 260 821.
We acknowledge VSB – Technical University of Ostrava, IT4Innovations National Supercomputing Center, Czech Republic, for awarding this project access to the LUMI supercomputer, owned by the EuroHPC Joint Undertaking, hosted by CSC (Finland) and the LUMI consortium through the Ministry of Education, Youth and Sports of the Czech Republic through the e-INFRA CZ (grant ID: 90254).

\vspace{-6pt}

\subsubsection{\discintname}
The authors have no competing interests to declare that are
relevant to the content of this article.
\end{credits}
\bibliographystyle{splncs04}
\bibliography{mybibliography}
\end{document}